\documentclass[10pt,journal,compsoc]{IEEEtran}

\ifCLASSOPTIONcompsoc
  \usepackage[nocompress]{cite}
\else
  \usepackage{cite}
\fi

\ifCLASSINFOpdf
\else
\fi

\usepackage{graphicx}
\usepackage[cmex10]{amsmath}
\usepackage{subfig}
\usepackage{array}
\usepackage{booktabs}
\usepackage{multirow}
\usepackage{threeparttable}
\usepackage{amsfonts}
\usepackage{tabularx}
\usepackage{algorithm}
\usepackage{algorithmic}
\usepackage{amsmath}
\usepackage{amssymb}
\DeclareMathOperator*{\argmax}{arg\,max}
\DeclareMathOperator*{\argmin}{arg\,min}
\usepackage{mathtools}
\usepackage{cuted}

\DeclarePairedDelimiter\abs{\lvert}{\rvert}%
\newcolumntype{Y}{>{\centering\arraybackslash}X}
\newcolumntype{b}{>{\centering\arraybackslash}X}
\newcolumntype{s}{>{\hsize=.33\hsize\centering\arraybackslash}X}

\makeatletter
\newcommand*{\rom}[1]{\expandafter\@slowromancap\romannumeral #1@}
\makeatother

\begin{document}
\title{Importance Weighted Structure Learning for Scene Graph Generation}

\author{Daqi~Liu,~Miroslaw~Bober,~\IEEEmembership{Member,~IEEE},~Josef~Kittler,~\IEEEmembership{Life Member,~IEEE}% <-this % stops a space
\IEEEcompsocitemizethanks{\IEEEcompsocthanksitem The authors are with the Centre for Vision, Speech and Signal Processing, University of Surrey, Guildford GU2 7XH, U.K.   \textbf{Under Review...}   \protect\\
E-mail: \{daqi.liu, m.bober, j.kittler\}@surrey.ac.uk}
}

%{Shell \MakeLowercase{\textit{et al.}}: Bare Demo of IEEEtran.cls for Computer Society Journals}

\IEEEtitleabstractindextext{%
\begin{abstract}
Scene graph generation is a structured prediction task aiming to explicitly model objects and their relationships via constructing a visually-grounded scene graph for an input image. Currently, the message passing neural network based mean field variational Bayesian methodology is the ubiquitous solution for such a task, in which the variational inference objective is often assumed to be the classical evidence lower bound. However, the variational approximation inferred from such loose objective generally underestimates the underlying posterior, which often leads to inferior generation performance. In this paper, we propose a novel importance weighted structure learning method aiming to approximate the underlying log-partition function with a tighter importance weighted lower bound, which is computed from multiple samples drawn from a reparameterizable Gumbel-Softmax sampler. A generic entropic mirror descent algorithm is applied to solve the resulting constrained variational inference task. The proposed method achieves the state-of-the-art performance on various popular scene graph generation benchmarks.
\end{abstract}

}

\maketitle

\IEEEdisplaynontitleabstractindextext

% \fi
\IEEEpeerreviewmaketitle

\IEEEraisesectionheading{\section{Introduction}\label{sec:introduction}}

\IEEEPARstart{A}{s} a structured prediction task, scene graph generation (SGG) aims to construct a visually-grounded scene graph for an input image, in which its potential objects as well as their relevant relationships are explicitly modelled. Such fundamental scene understanding task could potentially facilitate the downstream computer vision tasks, such as image captioning \cite{you2016image}, \cite{rennie2017self}, \cite{yang2019auto} and visual question answering \cite{teney2017graph}, \cite{anderson2018bottom}, \cite{shi2019explainable}. Given an input image $x$, SGG aims to infer the optimum interpretations $z^*$ by a max aposteriori (MAP) estimation $z^*=\argmax_{z} p_{\theta}(z|x)$, where $\theta$ is applied to parameterize the underlying posterior $p(z|x)$. Due to the exponential dependencies among the output variables, it is often computationally intractable to directly compute $p_{\theta}(z|x)$.

To this end, current SGG models generally follow a variational Bayesian (VB) \cite{wainwright2008graphical}, \cite{fox2012tutorial} framework, in which the variational inference step aims to approximate $p_{\theta}(z|x)$ with a computationally tractable variational distribution $q(z)$, while the variaitonal learning step tries to fit the underlying posterior $p_{\theta}(z|x)$ for the ground-truth training samples via a cross entropy loss. To estimate the optimum $q^*(z)$ and $\theta^*$, one needs to alternate the above variational inference and learning steps. For tractability, in current SGG models \cite{xu2017scene}, \cite{li2017scene}, \cite{dai2017detecting}, \cite{woo2018linknet}, \cite{yang2018graph}, \cite{wang2019exploring}, \cite{tang2019learning}, \cite{li2021bipartite}, the variational distribution $q(z)$ is often assumed to be fully decomposed as $q(z_1,z_2,...,z_n)=\prod_{i=1}^{n}q_i(z_i)$, where $z_i$ is the interpretation of one of the potential $n$ object and relationship region proposals and $q_i(z_i)$ represents the corresponding local variational approximation. The resulting VB framework is also known as the mean field variational Bayesian (MFVB) \cite{wainwright2008graphical}, \cite{fox2012tutorial}, and the associated variational inference step is also called mean field variational inference (MFVI) \cite{wainwright2008graphical}, \cite{fox2012tutorial}.

The above MFVI step in current SGG tasks is often formulated using message passing neural network (MPNN) models \cite{tang2019learning}, \cite{li2021bipartite}, \cite{li2018factorizable}, \cite{chen2019knowledge}, \cite{lin2020gps}, which require two fundamental modules to be constructed: visual perception and visual context reasoning \cite{liu2019visual}. Such formulation combines the superior feature representation learning capability of the deep neural networks and the inference capability of the classical MFVI. As a result, the above MPNN-based MFVI methodology has became the ubiquitous solution for current SGG tasks \cite{tang2019learning}, \cite{li2021bipartite}, \cite{li2018factorizable}, \cite{chen2019knowledge}, \cite{lin2020gps}. In the above formulation, a classical evidence lower bound (ELBO) \cite{zhang2018advances} is often implicitly (since the message passing optimization method do not need to explicitly maximize it) chosen as the variational inference objective.

However, the variational approximation inferred from such loose ELBO objective generally underestimates the underlying complex posterior \cite{zhang2018advances}, which often leads to inferior generation performance. In other words, the classical ELBO objective does not achieve a balanced bias-variance trade-off, since it generates overly simplified representations which fail to use the entire modeling capability of the network \cite{yuri2016importance}. This perhaps partly explains the fact that the detection performance of the current SGG models fall short of our expectations.

To solve the above issue, in this paper, we propose a novel importance weighted structure learning (IWSL) method, which employs a tighter importance weighted lower bound \cite{yuri2016importance} to replace the classical loose ELBO \cite{zhang2018advances} as the variational inference objective. Such importance weighted approximation is essentially a lower bound of the underlying log-partition function \cite{yuri2016importance}, which is estimated from the multiple samples drawn from a reparameterizable Gumbel-Softmax sampler \cite{eric2017categorical}, \cite{maddison2017concrete}. Unlike the classical MPNN-based SGG models, the proposed IWSL method requires to solve a constrained variational inference step. Basically, it aims to explicitly maximize the importance weighted lower bound objective, subject to the constraint that the categorical probability approximated by a Gumbel-Softmax variational distribution resides in a probability simplex. To this end, a generic constrained optimization algorithm - entropic mirror descent \cite{beck2003mirror} - is applied to infer the optimum interpretation from the input image. The proposed IWSL method achieves the state-of-the-art performance on two popular scene graph generation benchmarks: Visual Genome and Open Images V6.

This paper is organized as follows: Section 2 presents the related works while Section 3 demonstrates the proposed importance weighted structure learning methodology. The experimental results and the corresponding analysis are elaborated in Section 4. Finally, the conclusions are drawn in Section 5.

\section{Related Works}

Current SGG models follow two main research directions: pursuing a superior feature extracting structure or implementing an unbiased relationship prediction. The former aims to improve the feature representation learning capabilities of the neural network models, while the latter focuses on overcoming the problem of bias (which mainly detects the dominant relationship categories with abundant training samples and largely ignores the informative ones with fewer training samples) in the learnt relationship prediction,  caused by a long-tail data distribution.

For the first direction, \cite{yang2019auto}, \cite{dai2017detecting}, \cite{li2018factorizable}, \cite{qi2019attentive}, \cite{zellers2018neural} devise novel MPNN models while \cite{qi2019attentive}, \cite{yang2018graph}, \cite{tang2019learning}, \cite{woo2018linknet}, \cite{lin2020gps} embed the relevant contextual structural information into the current MPNN models. For the second direction, various debiasing methodologies have been proposed to solve the biased relationship prediction problem. For instance, dataset resampling \cite{chawla2002smote}, \cite{shen2016relay}, \cite{mahajan2018exploring}, instance-level resampling \cite{gupta2019lvis}, \cite{hu2020learning}, bi-level data resampling \cite{li2021bipartite}, knowledge transfer learning \cite{gidaris2018dynamic}, \cite{zhou2020bbn}, \cite{guo2021general} and loss reweighting based on instance frequency \cite{cao2019learning}, \cite{cui2019class} are the various remedies suggested in the literature. Unlike the above traditional debiasing methods, \cite{tang2020unbiased} removes the harmful bias from the good context bias based on the counterfactual causality (via calculating the Total Direct Effect with the help of a causal graph).

Most of the above SGG models \cite{zhang2019vrd}, \cite{zellers2018neural}, \cite{yang2018graph}, \cite{li2017scene}, \cite{lin2020gps}, \cite{li2021bipartite} tend to rely on a unified MPNN-based MFVB methodology. Such formulation generally employs the ELBO as the variational inference objective, in which the resulting variational approximation derived from ELBO often underestimates the underlying complex posterior \cite{yuri2016importance}. In contrary, we use a tighter importance weighted lower bound as the variational inference objective in the proposed IWSL method, and solve the resulting constrained variational inference task via a generic entropic mirror descent strategy rather than the traditional message passing technique. Specifically, various samples drawn from a reparameterizable Gumbel-Softmax sampler \cite{eric2017categorical}, \cite{maddison2017concrete} are applied to compute the above importance weighted lower bound. 

The above strictly tighter importance weighted lower bound is firstly introduced in the importance weighted autoencoder (IWAE) \cite{yuri2016importance}, which is a generative model with the same architecture as the classical variational autoencoder (VAE). In particular, the recognition network in IWAE relies on multiple samples to approximate the posterior, which increases the flexibility to model complex posteriors which do not fit the VAE modeling assumptions. Moreover, due to the inability to backpropagate through samples, the output categorical latent variables in SGG tasks are rarely employed in the stochastic neural networks. To this end, in this paper, instead of producing non-differentiable samples from a categorical distribution, Gumbel-Softmax sampler is utilized to draw differentiable samples from a novel Gumbel-Softmax distribution \cite{eric2017categorical}, \cite{maddison2017concrete}. Due to the applied explicit reparameterization function, it is quite easy to construct an efficient gradient estimator.

\section{Proposed Methodology}

In this section, we first present the problem formulation, followed by the proposed scoring function and the relevant Gumbel-Softmax sampler. Finally, the importance weighted structure learning and the adopted entropic mirror descent inference strategy are also discussed in the last two subsections. Fig.1 demonstrates the overview of the proposed IWSL method.
\begin{figure*}[!t]
\centering
\includegraphics[width= \linewidth]{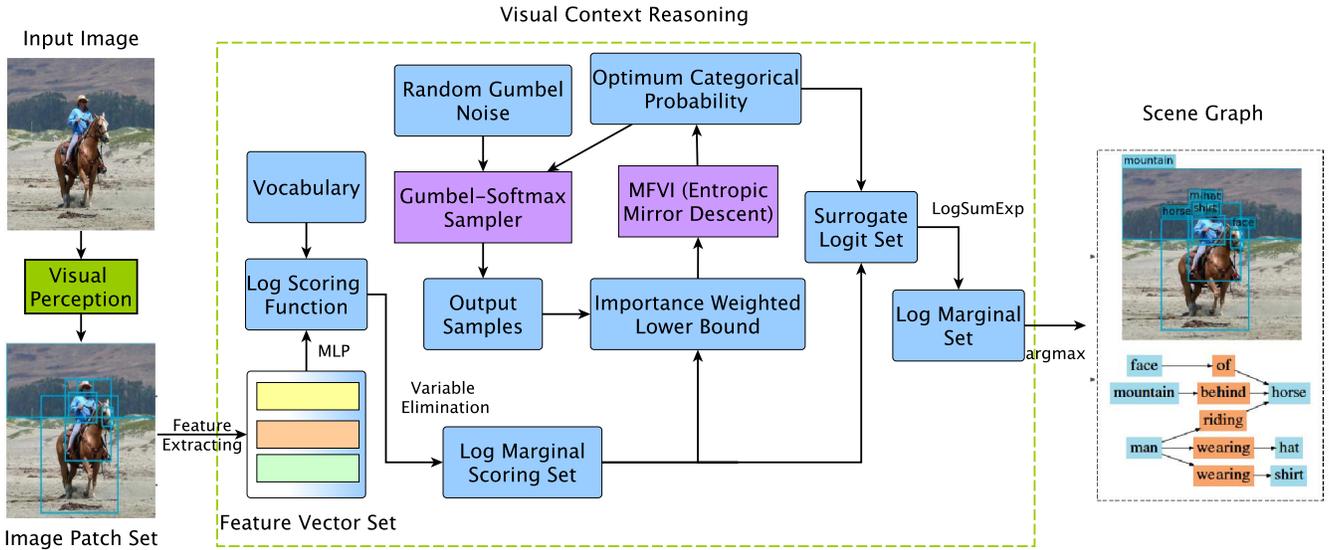}
\caption{ An overview of the proposed IWSL method. The green dash line designates the proposed visual context reasoning module. Given an input image, the visual perception module is used to generate a set of region proposals  with the associated image patches. The corresponding feature vectors are obtained by a certain feature extracting algorithm defined in \cite{ren2015faster}.  Given a feature vector set and a relevant vocabulary, the target log scoring function is computed via MLP. The variable elimination technique is applied to infer the corresponding log marginal scoring set, and the output samples are drawn from a Gumbel-Softmax sampler by feeding random Gumbel noises into it. With the log marginal scoring set and the above output samples, one can compute the corresponding importance weighted lower bound, and employ an MFVI framework (with the entropic mirror descent inference method) to infer the optimum categorical probability, which is then used to update the Gumbel-Softmax sampler. The associated surrogate logit set is computed from the above optimum categorical probability as well as the log marginal scoring set, which is then transformed into the resulting log marginal set via the $LogSumExp$ trick. Finally, the output scene graph is generated via the corresponding $\argmax$ operation. A cross-entropy loss is employed in the variational learning step.}
\label{fig_1}
\end{figure*}

\subsection{Problem Formulation}

As a structured prediction task, scene graph generation aims to build a visually-grounded scene graph for an input image by explicitly identifying the scene objects and their relevant relationships. In the current SGG settings, the scene graph includes a list of intertwined semantic triplet structures, in which each triplet consists of three components: a subject, a predicate and an object. Specifically, the current SGG tasks only focus on inferring the pairwise relationships, where the relationship between two interacting instances (subject and object) in an input image is termed as a predicate.

Generally, SGG task aims to infer the optimum interpretations $z^*$ from the input image $x$ via a MAP estimation $z^*=\argmax_{z}p_{\theta}(z|x)$, in which the underlying posterior $p_{\theta}(z|x)$ can be computed as follows:
\begin{equation}
 p_{\theta}(z|x)=\frac{s_{\theta}(x,z)}{\sum_{z}s_{\theta}(x,z)}=\frac{s_{\theta}(x,z)}{s_{\theta}(x)}
\end{equation}
where $s_{\theta}(x,z)$ is a scoring function of the probabilistic graphical model (e.g. conditional random field \cite{sutton2006introduction}) of the SGG task, which is used to measure the similarity/compatibility between the input variable $x$ and the output variable $z$. $s_{\theta}(x)$ is the relevant partition function or normalizing constant. Due to the exponential combinatorial dependencies existing in the SGG task, $s_{\theta}(x)$ is generally computationally intractable, which implies that it can only be estimated by certain approximation strategies. 

To this end, a classical variational inference (VI) \cite{wainwright2008graphical}, \cite{fox2012tutorial} technique is generally employed to estimate the computationally intractable log-partition function $logs_{\theta}(x)$ in current SGG models. For tractability, the computationally tractable variational distribution is often assumed to be fully decomposed and the resulting inference technique is known as mean field variational inference (MFVI) \cite{wainwright2008graphical}, \cite{fox2012tutorial}. Besides the above variational inference step, one still requires another variational learning step to fit the underlying posterior with the ground-truth training samples. Such formulation is also known as mean field variational Bayesian (MFVB) \cite{wainwright2008graphical}, \cite{fox2012tutorial} framework, in which the optimum $q^*(z)$ and $\theta^*$ are obtained by alternating the above inference and learning steps. More importantly, for MFVI, the original MAP inference in SGG task can be transformed into a corresponding marginal inference task, which is not the case for other VI techniques \cite{blake2011markov}. 

In the current SGG modelling approaches, a classical cross-entropy loss is generally employed to solve the variaitional learning step, while the above MFVI step is commonly formulated using a message passing neural network (MPNN) \cite{scarselli2008graph}, \cite{gilmer2017neural}, \cite{wang2018non}, \cite{zhou2020graph} model, in which the traditional ELBO is invariably implicitly %(since the message passing optimization strategy aims to implicitly maximize the ELBO) 
chosen as the variational inference objective. As a result, the MPNN-based MFVB formulation has became the de facto solution for current SGG tasks. In this MPNN-based MFVB framework, two fundamental modules are required, namely, visual perception and visual context reasoning. The former aims to locate and instantiate the objects and predicates within the input image, while the latter tries to infer their consistent interpretation. % according to certain inference strategies. 

Specifically, given an input image $x$, visual perception module aims to generate a set of object region proposals $b_i^o\in \mathbb{R}^4,i=1,...,m$, as well as a set of predicate region proposals $b_j^p\in \mathbb{R}^4,j=1,...,n$, where $m$ and $n$ represent the number of objects and predicates detected in the input image, respectively. Correspondingly, the input image $x$ can be divided into two sets of image patches $x_i^o,i=1,...,m$ and $x_j^p,j=1,...,n$. One can extract the relevant fixed-sized latent feature representation sets $y_i^o\in \mathbb{R}^d,i=1,...,m$ and $y_j^p \in \mathbb{R}^d, j=1,...,n$ by applying a ROI pooling on the feature maps obtained from the visual perception module. Given a set of object classes $\mathcal{C}$ and a set of relationship categories $\mathcal{R}$, a visual context reasoning module aims to infer the resulting object and predicate interpretation sets $z_i^o\in \mathcal{C},i=1,...,m$ and $z_j^p\in \mathcal{R},j=1,...,n$ based on the above latent feature representation sets. 

However, the variational distribution derived from the above loose ELBO objective often underestimates the underlying posterior \cite{zhang2018advances}, which leads to inferior generation performance. To this end, in this paper, we propose a novel importance weighted structure learning (IWSL) method, which employs a tighter (which is explicitly explained in Section 3.4) importance weighted lower bound to replace the classical ELBO as the variational inference objective. Instead of relying on the classical message passing technique, the proposed IWSL method applies a generic entropic mirror descent algorithm to accomplish the resulting constrained variational inference task. The proposed generic IWSL methodology allows us to increase the model complexity so that one can find a balanced bias-variance trade-off. In other words, the resulting variational distribution derived from the tighter importance weighted lower bound objective facilitates a better exploration of the complex multiple-modal behaviour of the underlying posterior.

\subsection{Scoring Function}

As a structured prediction task, scene graph generation can be generally formulated using a probabilistic graphical model, e.g. a conditional random field (CRF) \cite{sutton2006introduction}. With such probabilistic graphical model, one can model the conditional dependencies among the relevant variables by devising a non-negative scoring function $s_{\theta}(x,z)$, where $\theta$ is used to parameterize the scoring function. 

Basically, $s_{\theta}(x,z)$ measures the similarity or compatibility between the input image $x$ and the output interpretations $z$, which is often defined as follows:
\begin{equation}
 s_{\theta}(x,z)=\prod_{r\in R}f_r(x_r,z_r)
\end{equation}
where $r$ is a clique within a clique set $R$ (which is defined by the corresponding graph structure), $f_r$ is a non-negative factor function which models the dependencies between $x_r$ and $z_r$. Generally, the underlying posterior $p_{\theta}(z|x)$ related to $s_{\theta}(x,z)$ is often assumed to be a Gibbs distribution in traditional VB framework \cite{wainwright2008graphical}. Correspondingly, $f_r$ often takes the exponential form and the log scoring function is computed as follows:
\begin{equation}
 logs_{\theta}(x,z)=-\sum_{r\in R}\psi_r(x_r,z_r)
\end{equation}
where $\psi_r$ is also known as a potential function. In current SGG models, we have two types of potential functions: the unary potential function $\psi_u$ and the pairwise/binary potential function $\psi_b$.

Currently, only local contextual information is considered in most previous SGG models, while the global contextual information is largely ignored. In this paper, we aim to compute a latent global feature representation $y^g\in \mathbb{R}^d$ from the global region proposal $b^g$, where $b^g$ is obtained by the union of all the associated object/predicate region proposals in the input image. Correspondingly, $x^g$ is the relevant global image patch of $b^g$, and $z^g$ is its interpretation.

With the above definitions, by adding two types of pairwise potential terms $\psi_{b}^p(x_j^p, x^g,z_j^p, z^g)$ and $\psi_{b}^o(x_i^o, x^g,z_i^o, z^g)$, one could incorporate the global contextual information into the following applied log scoring function: 
\begin{equation}
\begin{split}
logs_{\theta}(x,z)= -\displaystyle\sum_{i=1}^{m}[\psi_u^o(x_i^o, z_i^o)+\displaystyle\sum_{j\in N(i)}\psi_{b}^o(x_i^o, x_j^p,z_i^o, z_j^p)\\
+\displaystyle\sum_{l\in N(i)}\psi_{b}^o(x_i^o, x_l^o,z_i^o,z_l^o)+\psi_{b}^o(x_i^o, x^g,z_i^o, z^g)]\\
-\displaystyle\sum_{j=1}^{n}[\psi_u^p(x_j^p, z_j^p)+
\displaystyle\sum_{i\in N(j)}\psi_{b}^p(x_i^o, x_j^p,z_i^o, z_j^p)\\
+\psi_{b}^p(x_j^p, x^g,z_j^p, z^g)] 
\end{split}
\end{equation}
where the superscripts $o$, $p$, $g$ represent the object, the predicate and the global context, respectively. $N(i)$ is the set of neighbouring nodes around the target $i$. It is worth to note the latent feature representations $y$ are implicitly embedded in the above formulation.

\subsection{Gumbel-Softmax Sampler}

In SGG tasks, the output variables $z$ are generally assumed as categorical variables, which are rarely applied in stochastic neural networks due to the inability to compute and backpropagate the gradients of the associated discrete distributions \cite{eric2017categorical}, \cite{maddison2017concrete}. To this end, instead of generating non-differentiable samples from a categorical distribution, a Gumbel-Softamx sampler \cite{eric2017categorical}, \cite{maddison2017concrete} is employed to produce differentiable samples drawn from a novel Gumbel-Softmax distribution. Such sampler is generally reparameterizable, in which an efficient gradient estimator can be easily implemented. 

Suppose $z$ be the interpretation of a potential region proposal, it can be modelled as a categorical variable with the class probabilities $\pi^1,..., \pi^v$ (where $v$ is the number of hypotheses for $z$), and the output categorical variables in SGG are encoded as $v$-dimensional one-hot vectors locating on the corners of the $(v-1)$-dimensional simplex, $\Delta^{v-1}$. The reparameterization function $g_{\pi}$ in Gumbel-Softmax sampler is defined as follows: 
\begin{equation}
    g_{\pi}(\sigma)=z
\end{equation}
where $\sigma$ represents a $v$-dimensional Gumbel noise, $z\in \Delta^{v-1}$ is the output $v$-dimensional sample vector and its $i$-th element $z^i$ is computed as follows:
\begin{equation}
 z^i=\frac{exp((log(\pi^i)+\sigma^i)/\tau)}{\sum_{j=1}^{v}exp((log(\pi^j)+\sigma^j)/\tau)},\; for\; i=1,...,v
\end{equation}
where $\tau$ is the softmax temperature. The samples drawn from the Gumbel-Softmax sampler become one-hot vectors when annealing the softmax temperature $\tau$ to zero. In reality, $\tau$ is often annealed to a relatively low temperature instead of zero. 

\subsection{Importance Weighted Structure Learning}

In traditional VB framework, ELBO is generally employed as the variational inference objective and the resulting maximization of ELBO is used to approximate the computationally intractable log-partition function \cite{zhang2018advances}. Given a scoring function $s_{\theta}(x,z)$ and a computationally tractable variational distribution $q(z)$, one can easily derive the following:
\begin{equation}
 logs_{\theta}(x)=\mathbb{E}_{q(z)}log\frac{s_{\theta}(x,z)}{q(z)}+\mathbb{E}_{q(z)}log\frac{q(z)}{p_{\theta}(z|x)}
\end{equation}
where, on the right-hand side of the equation, the first term is the so-called  ELBO and the second term is the Kullback–Leibler (KL) divergence between the variational distribution $q(z)$ and the underlying posterior $p_{\theta}(z|x)$. Clearly, ELBO is lower bound of $logs_{\theta}(x)$ since the KL divergence term is non-negative. However, such loose ELBO objective often leads to overly simplified model, which may not capture the complicated multi-modal structure of the underlying posterior \cite{yuri2016importance}. 

To this end, a tighter lower bound $\mathcal{L}_s$ based on $s$-sample importance weighting \cite{yuri2016importance} is employed to replace the classical ELBO in this paper. Such lower bound is also known as importance weighted lower bound, which is defined as follows:
\begin{equation}
  \mathcal{L}_{s}=\mathbb{E}_{z_1,...,z_s \sim q(z)}[log\frac{1}{s}\sum_{i=1}^{s}\frac{s_{\theta}(x,z_i)}{q(z_i)}] \leq logs_{\theta}(x)
\end{equation}
where $s$ represents the number of samples, in which each $z_i$ is an $i.i.d.$ random sample drawn from the variational distribution $q(z)$. $w_i=\frac{s_{\theta}(x,z_i)}{q(z_i)}$ is also known as the importance weight, and $\mathcal{L}_{s}$ is the importance weighted lower bound of $logs_{\theta}(x)$ when $s$ is a relatively small value. Essentially, $\mathcal{L}_{s}$ becomes an unbiased estimator of $logs_{\theta}(x)$ when $s$ reaches infinity. In particular, the traditional ELBO is just a special case of the importance weighted lower bound $\mathcal{L}_{s}$ (when setting $s=1$). Using more samples could only improve the tightness of the bound \cite{yuri2016importance}. Therefore, compared with the traditional ELBO, $\mathcal{L}_{s}$ is a much tighter lower bound of the log partition function $logs_{\theta}(x)$. 

For tractability, the above variational distribution $q(z)$ is often assumed to be fully decomposed as:
\begin{equation}
\begin{split}
q(z)=\displaystyle\prod_{i=1}^{m}q^o_i(z_i^o)\displaystyle\prod_{j=1}^{n}q^p_j(z_j^p)
\end{split}
\end{equation}
where $q^o_i(z_i^o) \in \Delta ^{v_o-1}$ and  $q^p_j(z_j^p) \in \Delta ^{v_p-1}$ are local variational approximations for the objects and predicates in the output scene graph, respectively. $v_o$ and $v_p$ are the sizes of vocabularies for the objects and predicates, respectively. Such inference procedure is also known as mean field variational inference (MFVI) \cite{wainwright2008graphical}, \cite{fox2012tutorial}. In MFVI, the MAP inference formulated in the first subsection can be transformed into a corresponding marginal inference task, which may not be the case in general \cite{blake2011markov}. 

Given a potential region proposal $b_i$, the corresponding local log marginal posterior $logp_{\theta}(z_i|x_i)$ is computed as follows:
\begin{equation}
  logp_{\theta}(z_i|x_i)=logs_{\theta}(x_i,z_i)-logs_{\theta}(x_i)
\end{equation}
where $logs_{\theta}(x_i)$ is the computationally intractable log partition function, and $logs_{\theta}(x_i,z_i)=\sum_{z\backslash z_i}logs_{\theta}(x_i,z)$ (where $z\backslash z_i$ represents marginalization over all nodes except the node $i$) is the local log marginal scoring function of $b_i$, which is generally obtained by a variable elimination technique. 

Specifically, for a potential object region proposal $b_i^o$, it is computed as follows:
\begin{equation}
\begin{split}
logs_{\theta}(x_i^o,z_i^o)\propto -[\psi_u^o(x_i^o, z_i^o)+\sum_{j\in N(i)} m^{op}_{j\to i}\\
+\sum_{l\in N(i)} m^{oo}_{l\to i}+m^{og}_{g\to i}] \\
\psi_u^o(x_i, z_i^o) = h^o_{\theta}(x_i)\cdot z_i^o\\
m^{op}_{j\to i}=\sum_{z_j^p\in \mathcal{R}}\psi^o_b(x_i^o, x_j^p, z_i^o, z_j^p)=g^{op}_{\theta}(x_i^o, x_j^p)\cdot z_i^o\\
m^{oo}_{l\to i}=\sum_{z_l^o\in \mathcal{C}}\psi^o_b(x_i^o, x_l^o, z_i^o, z_l^o)=g^{oo}_{\theta}(x_i^o, x_l^o)\cdot z_i^o\\
m^{og}_{g\to i}=\sum_{z^g\in \mathcal{G}}\psi^o_b(x_i^o, x^g, z_i^o, z^g)=g^{og}_{\theta}(x_i^o, x^g)\cdot z_i^o\\
\end{split}
\end{equation}
while for a potential predicate region proposal $b_j^p$, it is obtained by:
\begin{equation}
\begin{split}
logs_{\theta}(x_j^p,z_j^p)\propto -[\psi_u^p(x_j^p, z_j^p)+
\sum_{i\in N(j)} m^{po}_{i\to j}+m^{pg}_{g\to j}]\\
\psi_u^p(x_j, z_j^p) = h^p_{\theta}(x_j)\cdot z_j^p\\
m^{po}_{i\to j}=\sum_{z_i^o\in \mathcal{C}}\psi^p_b(x_i^o, x_j^p, z_i^o, z_j^p)=g^{po}_{\theta}(x_i^o, x_j^p)\cdot z_j^p\\
m^{pg}_{g\to j}=\sum_{z^g\in \mathcal{G}}\psi^p_b(x^g, x_j^p, z^g, z_j^p)=g^{pg}_{\theta}(x_j^p, x^g)\cdot z_j^p
\end{split}
\end{equation}
In the above two equations, $\cdot$ means an inner product, $z_i^o$ and $z_j^p$ are the output variables for an object and a predicate, which are generated by a Gumbel-Softmax sampler. $\mathcal{G}$ is a relevant global region proposal interpretation set. The feature representation learning functions $h^o_{\theta}$, $h^p_{\theta}$, $g^{op}_{\theta}$, $g^{oo}_{\theta}$, $g^{og}_{\theta}$, $g^{po}_{\theta}$, $g^{pg}_{\theta}$ are constructed by combing visual perception modules and multi-layer perceptrons (MLPs), which are parameterized by $\theta$. Each of these functions will first map the input image patches $x$ into the corresponding feature representations $y\in \mathbb{R}^d$ via the visual perception module, and then obtain the resulting $\mathbb{R}^v$ dimensional feature vector by feeding relevant $y$ into the corresponding MLP.  Most importantly, the MLPs implicitly perform the potential function marginalization prescribed in Equations (11) and (12). The resulting log score is essentially the inner product of the above $\mathbb{R}^v$ dimensional feature vector and the corresponding $v$-dimensional vector $z$.

To approximate the computationally intractable $logs_{\theta}(x_i)$, in this paper, we employ the following constrained variational inference objective $\mathcal{L}_s^i$:
\begin{equation}
\begin{split}
  &logs_{\theta}(x_i)\triangleq \max_{\pi_i}\mathcal{L}_s^i=\\
  &\max_{\pi_i}\mathbb{E}_{z_{i1},...,z_{is}\sim q_{\pi_i}(z_i)}[log\frac{1}{s}\sum_{j=1}^s\frac{s_{\theta}(x_i,z_{ij})}{q_{\pi_i}(z_{ij})}] \;
  s.t. \; \pi_i \in \Delta^{v-1}
\end{split}
\end{equation}
where $\mathcal{L}_s^i$ represents the $s$-sample importance weighted lower bound, and the local variational approximation $q_i(z_i)$ is set to a Gumbel-Softmax distribution with a categorical probability $\pi_i\in \Delta^{v-1}$. $z_{i1},...,z_{is}$ represent the $s$ $i.i.d$ samples drawn from $q_{\pi_i}(z_i)$. Since the local Gumbel-Softmax variational distribution $q_{\pi_i}(z_i)$ is reparameterizable, based on the Gumbel-Softmax sampler in the above subsection, $\mathcal{L}_s^i$ can be formulated as follows:
\begin{equation}
  \mathcal{L}_s^i=\mathbb{E}_{\sigma_{i1},...,\sigma_{is}\sim u(\sigma_i)}[log\frac{1}{s}\sum_{j=1}^s\frac{s_{\theta}(x_i,z_{ij})}{q_{\pi_i}(z_{ij})}]|_{z_{ij}=g_{\pi_i}(\sigma_{ij})}
\end{equation}
where $\sigma_{ij}$ is $v$-dimensional Gumbel noise drawn from a Gumbel distribution $u(\sigma_i)$, which is fed into the Gumbel-Softmax reparameterization function $g_{\pi_i}$ to explicitly compute the corresponding output sample $z_{ij}$.

As a result, the above expectation $\mathcal{L}_s^i$ can be approximated using a Monte Carlo estimator as follows:
\begin{equation}
  \mathcal{L}_s^i \triangleq [log\frac{1}{s}\sum_{j=1}^s\frac{s_{\theta}(x_i,z_{ij})}{q_{\pi_i}(z_{ij})}]|_{z_{ij}=g_{\pi_i}(\sigma_{ij}),\; \sigma_{ij \sim u(\sigma_i)}}
\end{equation}
where the efficient computation of the log importance weight $logw_{ij}=logs_{\theta}(x_i,z_{ij})-logq_{\pi_i}(z_{ij})$ is essential for computing the above $\mathcal{L}_s^i$. Specifically, $logs_{\theta}(x_i,z_{ij})$ can be easily computed based on Equation (11) and (12), while the above log probability $logq_{\pi_i}(z_{ij})$ is approximated as follows:
\begin{equation}
 logq_{\pi_i}(z_{ij})\triangleq \lVert \pi_i\cdot z_{ij} \rVert_1 -max(\pi_i)-log \lVert e^{\pi_i-max(\pi_i)}\rVert_1
\end{equation}
where $\lVert. \rVert_1$ represents the $\mathbb{L}_1$ norm while $max(\pi_i)$ is the maximum value of $\pi_i$. 

With the above $\mathcal{L}_s^i$, one can compute the target log marginal posterior $logp_{\theta}(z_i|x_i)$ via a corresponding surrogate logit $\phi$:
\begin{equation}
 \begin{split}
     logp_{\theta}(z_i|x_i)\triangleq \phi+C\\
     \phi=logs_{\theta}(x_i,z_i)-\max_{\pi_i}\mathcal{L}_s^i
 \end{split}
\end{equation}
where $C$ is a relevant constant w.r.t. $x_i$ and $z_i$. According to the $LogSumExp$ trick, one can compute $logp_{\theta}(z_i|x_i)$ by ignoring the above constant $C$:
\begin{equation}
\begin{split}
logp_{\theta}(z_i|x_i)\triangleq \phi - log{\lVert e^{\phi} \rVert_1}
\end{split}
\end{equation}
where the optimum interpretation $z_i^*$ of the input region proposal $b_i$ is computed as $z_i^*=\argmax_{z_i}logp_{\theta}(z_i|x_i)$. 

Moreover, a classical cross-entropy loss is employed in the relevant variational learning step to fit the above $p_{\theta}(z|x)$ with the ground-truth training samples:
\begin{equation}
    \theta^*=\argmin_{\theta}\mathbb{L}(\theta)=\argmin_{\theta}-[\frac{1}{c}\sum_{k=1}^{c}logp_{\theta}(\hat{z_k}|\hat{x_k})]
\end{equation}
where $\mathbb{L}(\theta)$ represents the variational learning objective, $c$ is the number of training images in a mini-batch, $\hat{z_k}$ is the ground-truth scene graph of the input image $\hat{x_k}$. 
\begin{algorithm}[ht]
 \caption{Importance Weighted Structure Learning}\label{euclid1}
 \textbf{Input} region proposal $b$, categorical probability $\pi$, number of samples $s$, Gumbel noise distribution $u(\sigma)$, Gumbel-Softmax reparameterization function $g_{\pi}$, learning rate $\alpha$, softmax temperature $\tau$, minimum temperature $\tau_{min}$, temperature annealing rate $\beta$, number of iterations $T$ \\
 \textbf{Output} $\theta$, $\tau$
 \begin{algorithmic}[1]
 \STATE randomly initialize $\theta$ 
 \FOR{iteration $t=1$ to $T$}
 \STATE randomly initialize $\pi$ for $b$
 \STATE draw $s$ Gumbel noise samples $\sigma_{1},...,\sigma_{s}$ from $u(\sigma)$
 \STATE compute $s$ output samples $z_{1},...,z_{s}$ by feeding $\sigma_{1},...,\sigma_{s}$ into $g_{\pi}$
 \STATE compute log importance weight $log\frac{s_{\theta}(x,z)}{q_{\pi}(z)}$ and approximate $\mathcal{L}_s$ via Monte Carlo estimation
 \STATE employ EMD to solve Equation (13) and update $\pi$
 \STATE compute the surrogate logit $\phi$ as well as the resulting $logp_{\theta}(z|x)$ using the updated $\pi$
 \STATE compute $\mathbb{L}(\theta)$ and update $\theta \leftarrow \theta - \alpha \cdot \bigtriangledown_{\theta} \mathbb{L}(\theta)$
 \STATE update $\tau \leftarrow \max(\tau \cdot e^{-\beta \cdot t}, \tau_{min})$
 \ENDFOR
 \end{algorithmic}
 \end{algorithm}

Finally, to better illustrate the proposed IWSL method, we summarize it in Algorithm 1. As a MFVB framework, the proposed IWSL method consists of two procedures: variational inference and variational learning. In particular, the variational inference procedure includes steps 3-7, while steps 8-9 represent the variational learning procedure. The temperature annealing process is accomplished in step 10. For computational efficiency, the number of samples $s$ in variational inference is often smaller than the one applied in variational learning.

Specifically, in variational inference, one first randomly initialize a categorical probability $\pi$ for a potential region proposal $b$, and then draw $s$ Gumbel noise samples $\sigma_{1},...,\sigma_{s}$ from a Gumbel noise distribution $u(\sigma)$. The output samples $z_{1},...,z_{s}$ are explicitly computed by feeding the above Gumbel noises $\sigma_{1},...,\sigma_{s}$ into a Gumbel-Softmax reparameterization function $g_{\pi}$. With the Equations (11), (12) and (16), one can easily compute the relevant importance weight $log\frac{s_{\theta}(x,z)}{q_{\pi}(z)}$ and approximate the corresponding importance weighted lower bound $\mathcal{L}_s$ via an Monte Carlo estimation. Finally, an entropic mirror descent (EMD) (which is introduced in the following subsection) method is applied to solve the resulting constrained variational inference task and obtain the optimum categorical probability $\pi^*$. In variational learning, according to Equation (17), one can first compute a surrogate logit $\phi$ based on the above updated optimum categorical probability $\pi^*$, and then obtain the resulting $logp_{\theta}(z|x)$ using the $LogSumExp$ trick. Finally, we compute the variational learning objective $\mathbb{L}(\theta)$ and then update the corresponding $\theta$ according to the stochastic gradient descent method.

With the above variational inference and learning procedures, in the training period, one can obtain the optimum $\theta^*$ and $\tau^*$. In the test period, $\theta$ and $\tau$ are fixed to $\theta^*$ and $\tau^*$, respectively. In particular, one discards the steps 9 and 10 in Algorithm 1, and generates the optimum interpretations $z^*=\argmax_{z}logp_{\theta}(z|x)$.

\subsection{Entropic Mirror Descent}

As demonstrated in Equation (13), the variational inference procedure in the proposed method requires us to solve a constrained optimization problem. Specifically, to approximate the computationally intractable log partition function $logs_{\theta}(x_i)$, one needs to maximize the $s$-sample importance weighted lower bound $\mathcal{L}_s^i$, subject to the constraint that the categorical probability $\pi_i$ resides in $(v-1)$-simplex.
\begin{algorithm}[ht]
 \caption{Entropic Mirror Descent}\label{euclid2}
 \textbf{Input} variational distribution $\pi$, importance weighted lower bound $\mathcal{L}_s$, number of iterations $M$, an initial learning rate $\gamma$, a predefined objective $\mathcal{L}_s^p$, a small positive value $\epsilon$\\
 \textbf{Output} optimum $\pi^*$
 \begin{algorithmic}[1]
 \FOR{iteration $i=1$ to $M$}
 \STATE compute the derivative $\bigtriangledown_{\pi}\mathcal{L}_s$
 \STATE set learning rate $\gamma = \frac{\gamma}{\sqrt{i}}$
 \STATE end the loop if {$\abs{\mathcal{L}_s-\mathcal{L}_s^p}<\epsilon$}
 \STATE set $\mathcal{L}_s^p=\mathcal{L}_s$
 \STATE compute $r=\gamma\cdot\bigtriangledown_{\pi}\mathcal{L}_s$
 \STATE compute $r=\pi\cdot e^{r-\max(r)}$
 \STATE set $\pi=\frac{r}{\lVert r \rVert_1}$
 \ENDFOR
 \end{algorithmic}
 \end{algorithm}

Among the existing constrained optimization algorithms, entropic mirror descent (EMD) \cite{beck2003mirror} method is chosen as the applied variational inference methodology, as demonstrated in Algorithm 2. Since the above constraint is a probability simplex, the negative entropy can naturally be employed as a specific function to construct the required Bregman distance \cite{teboulle1992entropic}. Due to the utilization of the geometry of the optimization problem \cite{raskutti2015information}, EMD generally converges faster than the classical projected gradient descent methods \cite{eicke1992iteration}.
\begin{table*}[t]
   \begin{threeparttable}
	\renewcommand{\arraystretch}{1.5}
	\caption{A performance comparison on Visual Genome dataset.}
    \centering
    \begin{tabular*}{\linewidth}{c|@{\extracolsep{\fill}}cccccc|cccc}
	\toprule
	\multirow{2}{*}{Method} & \multicolumn{2}{c}{PredCls} & \multicolumn{2}{c}{SGCls} & \multicolumn{2}{c|}{SGDet} & \multicolumn{4}{c}{SGDet(R@100)}\\ \cmidrule{2-3} \cmidrule{4-5} \cmidrule{6-7} \cmidrule{8-11} 
	 {} & {mR@50} & {mR@100} & {mR@50} & {mR@100} & {mR@50} & {mR@100} & {Head} & {Body} & {Tail} & {Mean}\\
	\midrule
    RelDN$ ^{\dagger}$\cite{zhang2019vrd}  & $15.8$ & $17.2$ & $9.3$ & $9.6$ & $6.0$ & $7.3$ & $34.1$ & $6.6$ & $1.1$ & $13.9$ \\
	Motifs\cite{zellers2018neural}  & $14.6$ & $15.8$ & $8.0$ & $8.5$ & $5.5$ & $6.8$ & $36.1$ & $7.0$ & $0.0$ & $14.4$\\
	Motifs*\cite{zellers2018neural}  & $18.5$ & $20.0$ & $11.1$ & $11.8$ & $8.2$ & $9.7$ & $34.2$ & $8.6$ & $2.1$ & $15.0$\\
	G-RCNN$ ^{\dagger}$\cite{yang2018graph}   & $16.4$ & $17.2$ & $9.0$ & $9.5$ & $5.8$ & $6.6$ & $28.6$ & $6.5$ & $0.1$ & $11.7$\\
	MSDN$ ^{\dagger}$\cite{li2017scene}   & $15.9$ & $17.5$ & $9.3$ & $9.7$ & $6.1$ & $7.2$ & $35.1$ & $5.5$ & $0.0$ & $13.5$\\
    GPS-Net$ ^{\dagger}$\cite{lin2020gps}   & $15.2$ & $16.6$ & $8.5$ & $9.1$ & $6.7$ & $8.6$& $34.5$ & $7.0$ & $1.0$ & $14.2$\\
    GPS-Net$ ^{\dagger *}$\cite{lin2020gps}   & $19.2$ & $21.4$ & $11.7$ & $12.5$ & $7.4$ & $9.5$& $30.4$ & $8.5$ & $3.8$ & $14.2$\\
    VCTree-TDE\cite{tang2020unbiased}  & $25.4$ & $28.7$ & $12.2$ & $14.0$ & $9.3$ & $11.1$ & $24.5$ & $13.9$ & $0.1$ & $12.8$\\
    BGNN\cite{li2021bipartite}  & $30.4$ & $32.9$ & $14.3$ & $16.5$ & $10.7$ & $12.6$& $33.4$ & $13.4$ & $6.4$ & $17.7$\\
    \textbf{IWSL} & $\mathbf{30.0}$ & $\mathbf{32.1}$ & $\mathbf{17.4}$ & $\mathbf{18.9}$ & $\mathbf{13.7}$ & $\mathbf{15.9}$& $\mathbf{30.6}$ & $\mathbf{16.5}$ & $\mathbf{10.7}$ & $\mathbf{19.3}$\\
	\bottomrule
    \end{tabular*}
    \begin{tablenotes}
	\item [\textbullet] Note: All the above methods apply ResNeXt-101-FPN as the backbone. $*$ means the re-sampling strategy \cite{gupta2019lvis} is applied in this method, and $\dagger$ depicts the reproduced results with the latest code from the authors. Using bold to represent the proposed method.
      \end{tablenotes}
    \end{threeparttable}
\end{table*} 

\section{Experiments}

In this section, we first validate the proposed IWSL method by comparing it with various state-of-the-art SGG models on two popular benchmarks: Visual Genome \cite{krishna2017visual} and Open Images V6 \cite{alina2020open}, respectively. Finally, the ablation study and the visualization results are presented and discussed in the last two subsections. 

\subsection{Visual Genome}
\subsubsection{Experiment Configuration}
\textbf{Benchmark:} Visual Genome \cite{krishna2017visual} is the most common scene graph generation benchmark, which consists of 108,077 images with an average of 38 objects and 22 relationships per image. In this experiment, we employ the data split protocol as in \cite{xu2017scene}, in which the most frequent 150 object categories and 50 predicate classes are selected. Furthermore, we split the Visual Genome into a training set ($70\%$) and a test set ($30\%$). For validation, an evaluation set ($5k$) is randomly selected from the training set. Following \cite{liu2019large}, according to the number of objects in training split, the relevant categories are divided into three disjoint sets: $head$ (more than $10k$), $body$ ($0.5k\sim 10k$) and $tail$ (less than $0.5k$), as demonstrated in Fig.2.

\noindent \textbf{Evaluation Metrics:} Due to the reporting bias caused by the data imbalance, the mean Recall$@K$ ($mR@K$) rather than the common Recall$@K$ ($R@K$) is chosen as the evaluation metric in this experiment. Compared with $R@K$ which only concentrates on common predicates (e.g. $on$) with abundant training samples, $mR@K$ focuses on the informative predicate categories (e.g. $parked\; on$) with much less training samples. Three tasks are applied to validate the proposed method, namely, Predicate Classification (PredCls), Scene Graph Classification (SGCls) and Scene Graph Detection (SGDet). Specifically, PredCls aims to predict the predicate labels, given the input image, the ground-truth bounding boxes and object labels; SGCls tries to predict the labels for objects and predicates, given the input image and the ground-truth bounding boxes; SGDet constructs the output scene graph from the input image. 
\begin{figure}[!t]
\centering
\includegraphics[width=\linewidth]{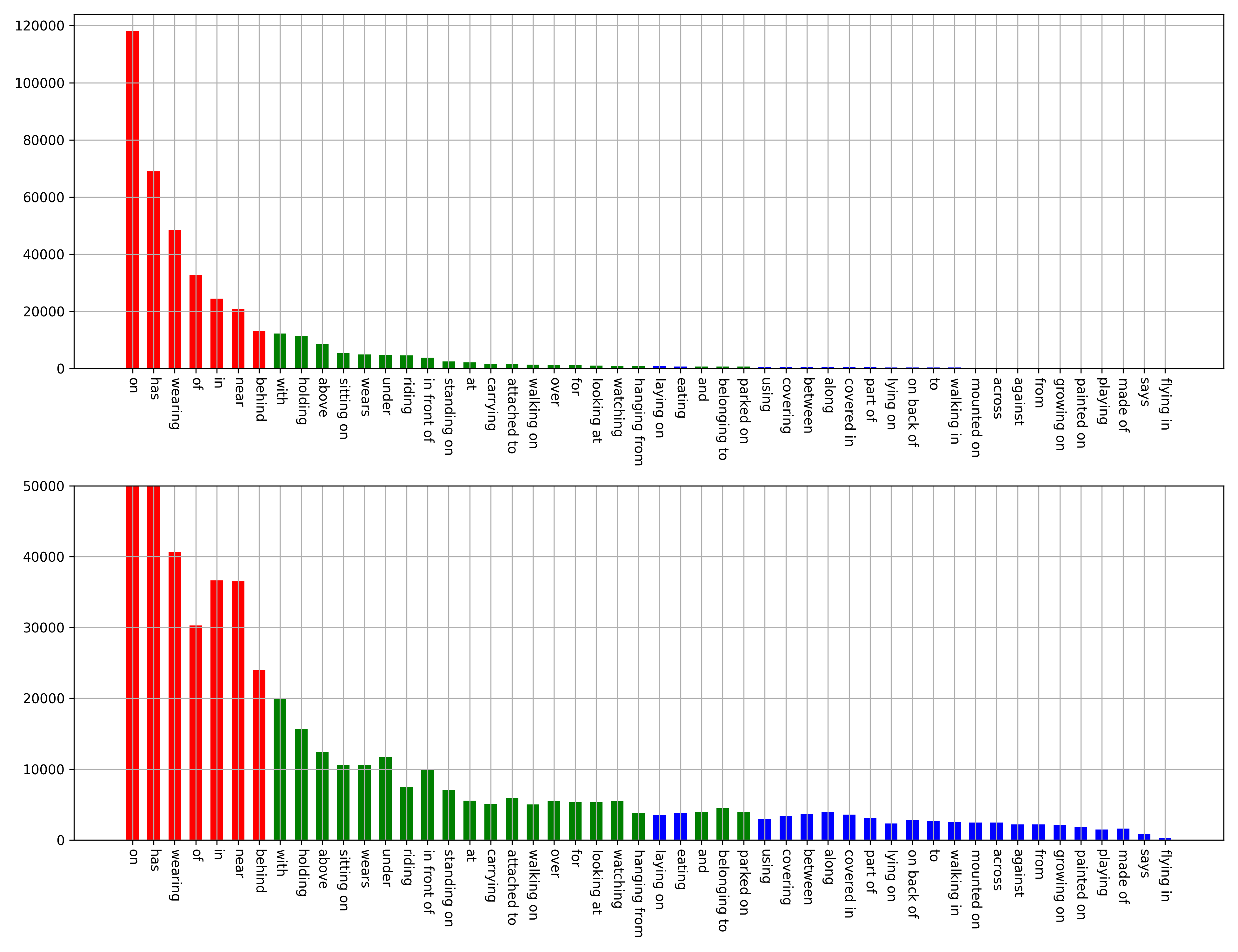}
\caption{Visual Genome training split follows a long-tail data distribution, which can be divided into three disjoint sets: $head$ (red bars), $body$ (green bars) and $tail$ (blue bars). Here, $y$ axis represents the number of samples.}
\label{fig_2}
\end{figure} 

\noindent \textbf{Implementation Details:} Following \cite{tang2020unbiased}, ResNeXt-101-FPN \cite{he2016deep} and Faster-RCNN \cite{ren2015faster} are chosen as the backbone and the object detector, respectively. Like previous methods, we choose the step training strategy and freeze the above visual perception models during training. As in \cite{li2021bipartite}, a bi-level data resampling strategy is adopted to achieve an effective trade-off between the head and the tail categories, which consists of image-level over-sampling and instance-level under-sampling. The former creates a random permutation of images, repeating each image according to its repeat factor $t$ in each epoch, while the latter achieves under-sampling according to a drop-out probability for instances of different predicate classes in each image. Specifically, we set the repeat factor $t=0.07$ and the instance drop rate $\gamma_{d}=0.7$. The batch size $bs$ is set to 12, and an SGD optimizer with a learning rate of $0.008\times bs$ is applied in this experiment. The number of samples $s$ is set to $50$ in the variational inference step. For the variational learning step, $s$ is set to $8000$ in the SGDet task and $5000$ in the PredCls and SGCls tasks.
\begin{table}[!t]
   \resizebox{\columnwidth}{!}{
   \begin{threeparttable}
	\renewcommand{\arraystretch}{1.5}
	\caption{The performance comparison on the Visual Genome dataset using the balance adjustment strategy.}
	\centering
    \begin{tabular}{@{\extracolsep{4pt}}*7c@{}}
	\toprule
	{} & \multicolumn{2}{c}{PredCls} & \multicolumn{2}{c}{SGCls} & \multicolumn{2}{c}{SGDet}\\ \cmidrule{2-3} \cmidrule{4-5} \cmidrule{6-7}
	{Method} & {mR@50} & {mR@100} & {mR@50} & {mR@100} & {mR@50} & {mR@100}\\
	\midrule
	Motifs+BA\cite{guo2021general}  & $29.7$ & $31.7$ & $16.5$ & $17.5$ & $13.5$ & $15.6$\\
	VCTree+BA\cite{guo2021general}  & $30.6$ & $32.6$ & $20.1$ & $21.2$ & $13.5$ & $15.7$\\
    Transformer+BA\cite{guo2021general}   & $31.9$ & $34.2$ & $18.5$ & $19.4$ & $14.8$ & $17.1$\\
    \textbf{IWSL+BA} & $\mathbf{36.9}$ & $\mathbf{39.2}$ & $\mathbf{20.7}$ & $\mathbf{22.2}$ & $\mathbf{16.1}$ & $\mathbf{18.6}$\\
	\bottomrule
    \end{tabular}
    \begin{tablenotes}
	\item [\textbullet] Note: All the above methods apply the same balance adjustment strategy as in \cite{guo2021general}. Using bold to represent the proposed method.
      \end{tablenotes}
    \end{threeparttable}
    }
\end{table} 

\subsubsection{Comparisons with State-of-the-art Methods}

As demonstrated in Table 1, besides achieving comparable performance with the latest BGNN algorithm in the PredCls task, the proposed IWSL method outperforms the previous state-of-the-art SGG models by a large margin in the remaining SGCls and SGDet tasks. For instance, for the most difficult yet representative SGDet task, compared with the latest BGNN model, the proposed IWSL method achieves $28\%$ and $26.2\%$ performance gain, respectively. Moreover, such performance is obtained with a quite small number of training iterations (which is usually set to $4000$). This is mainly because the applied generic entropic mirror descent method converges faster than the classical message passing strategy.

Moreover, we also compare the SGDet performance ($R@100$) on the long-tail categorical groups in Table 1, in which the proposed IWSL method archives the best mean performance. For the informative $body$ and $tail$ category groups, the proposed IWSL method outperforms the state-of-the-art SGG models by a large margin. Unlike the previous models, which focus on detecting the common predicate categories, the proposed IWSL method aims to improve the informative predicate detection capability. In other words, the proposed IWSL method could mitigate the intrinsic biased predicate prediction problem caused by the long-tail data distribution exhibited in Visual Genome. 

Generally, two types of imbalance lead to the above biased predicate prediction problem, namely the semantic space imbalance and the training sample imbalance. By adopting a generic balance adjustment (BA) strategy \cite{guo2021general} in the proposed IWSL method, the resulting IWSL+BA algorithm improves the informative predicate detection capability further. Specifically, to overcome the above two types of imbalance, two procedures are devised in the generic balance adjustment strategy: semantic adjustment and balanced predicate learning. The former aims to induce the predictions by the IWSL method to be more informative via constructing an appropriate transition matrix, while the latter tries to extend the sampling space for informative predicates%(in reference to the Shannon information theory, where the predicates occurring less frequently are deemed to contain more information)
.

As demonstrated in Table 2, for a fair comparison, the proposed IWSL+BA method is compared with three baseline models as presented in \cite{guo2021general}. It can be seen that the proposed IWSL+BA method outperforms the previous state-of-the-art models by a large margin, especially for the PredCls task. This is mainly thanks to: 1) the transition matrix, introduced by the semantic adjustment procedure, which maps the predictions from the IWSL method to more informative ones; 2) more balanced training samples drawn by the balanced predicate learning procedure, which discards some redundant training samples of the common $head$ group, while keeping the training samples in the informative $body$ and $tail$ groups.

\subsection{Open Images V6}
\subsubsection{Experiment Configuration}
\textbf{Benchmark:} With a superior annotation quality, Open Images V6 \cite{alina2020open} is another popular scene graph generation benchmark, which is constructed by Google. It includes 126,368 training images, 5322 test images and 1813 validation images. In this experiment, we adopt the same data processing protocols as in \cite{lin2020gps}, \cite{zhang2019vrd}, \cite{alina2020open}.

\noindent\textbf{Evaluation Metrics:} According to the evaluation protocols in \cite{lin2020gps}, \cite{zhang2019vrd}, \cite{alina2020open}, in this experiment, we choose the following evaluation metrics: the mean Recall$@50$ ($mR@50$), the regular Recall$@50$ ($R@50$), the weighted mean AP of relationships ($wmAP_{rel}$) and the weighted mean AP of phrase ($wmAP_{phr}$). Like \cite{lin2020gps}, \cite{alina2020open}, \cite{zhang2019vrd}, the weight metric score is defined as: $score_{wtd}=0.2\times R@50 + 0.4\times wmAP_{rel} + 0.4\times wmAP_{phr}$.
\begin{figure*}[!t]
\centering
\includegraphics[width= \linewidth]{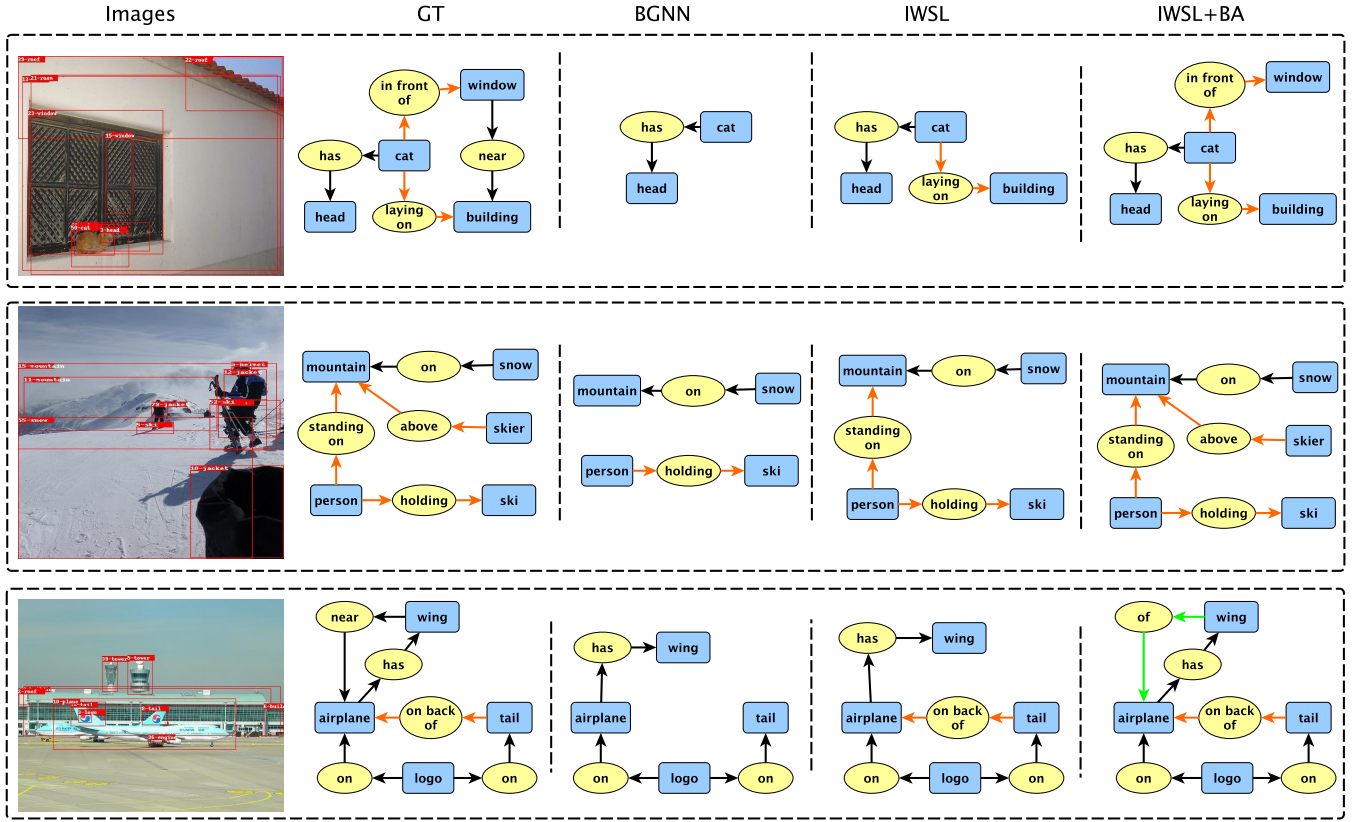}
\caption{Visualization of the qualitative results of the ground-truth (GT), the baseline BGNN model, the proposed IWSL method and the derived IWSL+BA algorithm in the SGDet task. The black, orange and green arrows represent the triplets with $head$ predicate categories, the triplets with $body$ or $tail$ predicate categories and the reasonable triplets detected by models which are not included in GT, respectively. Compared with the baseline BGNN model, the scene graphs generated by the proposed IWSL method and the derived IWSL+BA algorithm are much closer to the ground-truth scene graph GT.}
\label{fig_3}
\end{figure*}

\noindent \textbf{Implementation Details:} Similar to the previous experiment in Visual Genome, we choose ResNeXt-101-FPN \cite{he2016deep} and Faster-RCNN \cite{ren2015faster} as the backbone and the object detector, respectively. We employ the step training strategy and freeze the above visual perception models. The same bi-level data resampling strategy is adopted as in the previous experiment. We set the batch size $bs=12$ and utilize an Adam optimizer with the learning rate of $0.0001$. For the variational inference step, the number of samples $s$ is set to $50$. For the variational learning step, $s$ is set to $5000$.

\subsubsection{Comparisons with State-of-the-art Methods}
To verify the effectiveness of the proposed IWSL method further, in this experiment, we compare it with various state-of-the-art SGG models in Table 3. Some of the methods (with $\dagger$) are reproduced with the latest code from the authors, while the other methods (with $*$) employ an additional re-sampling strategy \cite{gupta2019lvis}. As demonstrated in Table 3, the proposed IWSL method outperforms the state-of-the-art SGG models by a large margin on the most representative $mR@50$ evaluation metric, and achieve a comparable performance with the latest BGNN algorithm on the remaining evaluation metrics.
\begin{table}[!t]
   \resizebox{\columnwidth}{!}{
   \begin{threeparttable}
	\renewcommand{\arraystretch}{1.5}
	\caption{A performance comparison on the Open Images V6 dataset.}
	\centering
    \begin{tabular}{@{\extracolsep{4pt}}*6c@{}}
	\toprule
	{Method} & {mR@50} & {R@50} & {wmAP\_rel} & {wmAP\_phr} & {score\_wtd} \\
	\midrule
	RelDN$^{ \dagger}$\cite{zhang2019vrd}  & $33.98$ & $73.08$ & $32.16$ & $33.39$ & $40.84$\\
	RelDN$^{\dagger*}$\cite{zhang2019vrd}  & $37.20$ & $75.34$ & $33.21$ & $34.31$ & $41.97$ \\
    VCTree$^{\dagger}$\cite{tang2019learning} & $33.91$ & $74.08$ & $34.16$ & $33.11$ & $40.21$ \\
    G-RCNN$^{\dagger}$\cite{yang2018graph}   & $34.04$ & $74.51$ & $33.15$ & $34.21$ & $41.84$ \\
	Motifs$^{\dagger}$\cite{zellers2018neural}  & $32.68$ & $71.63$ & $29.91$ & $31.59$ & $38.93$ \\
    VCTree-TDE$^{\dagger}$\cite{tang2020unbiased}  & $35.47$ & $69.30$ & $30.74$ & $32.80$ & $39.27$ \\
    GPS-Net$^{\dagger}$\cite{lin2020gps}   & $35.26$ & $74.81$ & $32.85$ & $33.98$ & $41.69$ \\
    GPS-Net$^{\dagger *}$\cite{lin2020gps}   & $38.93$ & $74.74$ & $32.77$ & $33.87$ & $41.60$ \\
    BGNN\cite{li2021bipartite}  & $40.45$ & $74.98$ & $33.51$ & $34.15$ & $42.06$ \\
    \textbf{IWSL} & $\mathbf{42.18}$ & $\mathbf{74.68}$ & $\mathbf{33.11}$ & $\mathbf{34.33}$ & $\mathbf{41.87}$ \\
	\bottomrule
    \end{tabular}
    \begin{tablenotes}
	\item [\textbullet] Note: All the above methods apply ResNeXt-101-FPN as the backbone. $*$ means the re-sampling strategy \cite{gupta2019lvis} is applied in this method, and $\dagger$ depicts the reproduced results with the latest code from the authors. Using bold to represent the proposed method.
      \end{tablenotes}
    \end{threeparttable}
    }
\end{table}

\subsection{Ablation Study}
The importance weighted lower bound becomes an unbiased estimator when the number of samples reaches infinity. However, it is impossible to achieve the above learning scenario in reality, due to the high computational complexity and the huge computational resources. In practice, for computational efficiency, the number of samples is often set to a relatively small number. To investigate the detection performance dependency of the proposed IWSL method on the number of samples, in this section, we choose three settings and compare their impact on the SGDet performance as shown in Table 4. We observe that the SGDet performance gradually increases with the number of samples. The performance gain obtained from a larger number of samples ($s>50$) is not that obvious, while its corresponding computation time increases dramatically. To pursue a balanced trade-off between the detection performance and the computational complexity, in this paper, the number of samples $s$ is set to $50$ in variational inference step. 
\begin{table}[!t]
   \begin{threeparttable}
	\renewcommand{\arraystretch}{1.5}
	\caption{Ablation study of the impact of the number of samples in variational inference step.}
	\centering
    \begin{tabularx}{\linewidth}{bsss}
	\toprule
	{Number of Samples $S$} &  {mR@20} & {mR@50} & {mR@100}\\ 
	\midrule
    $10$ & $9.9$ & $13.1$ & $15.5$ \\
	$30$ & $10.7$ & $13.5$ & $15.6$ \\
    $50$ & $10.6$ & $13.7$ & $15.9$\\
	\bottomrule
    \end{tabularx}
    \begin{tablenotes}
	\item [\textbullet] Note: We compare the SGDet performance in this ablation study.
      \end{tablenotes}
    \end{threeparttable}
\end{table} 

\subsection{Visualization Results}

To visually demonstrate the superiority of the proposed methods, in Fig. 3, we compare the visualization of the qualitative results of the ground-truth (GT), the baseline BGNN model, the proposed IWSL method and the derived IWSL+BA algorithm in the SGDet task. Compared with the baseline BGNN model, the proposed IWSL method is capable of detecting more informative $body$ and $tail$ predicates. For example, the proposed IWSL could detect an additional $<laying\; on>$ predicate for the top image, or an additional $<standing\; on>$ predicate for the middle image. Besides, the spatial informative predicates such as $<on\; back\; of>$ can also be detected. Moreover, the derived IWSL+BA method further improves the above capability. For example, it could detect an additional triplet $<cat\; in\; front\; of\; window>$ for the top image, or even a new reasonable triplet $<wing\; of\; airplane>$ (which is not included in the ground-truth scene graph GT) for the bottom image. In a word, compared with the baseline BGNN model, the scene graphs generated by the proposed IWSL method and the derived IWSL+BA algorithm are much closer to the ground-truth scene graph GT.

\section{Conclusion}

To achieve a balanced bias-variance trade-off, in this paper, we propose a novel importance weighted structure learning (IWSL) method, which employs a tighter importance weighted lower bound to replace the classical ELBO as the variational inference objective. This is because the variational approximation derived from the ELBO often underestimates the underlying posterior. A generic entropic mirror descent algorithm, rather than the traditional message passing strategy, is employed to accomplish the resulting constrained variational inference task. We validate the proposed IWSL method on two popular scene graph generation benchmarks: Visual Genome and Open Images V6, showing it outperforms the state-of-the-art models by a large margin.

\ifCLASSOPTIONcompsoc
  \section*{Acknowledgments}
\else
  \section*{Acknowledgment}
\fi

This work was supported in part by the U.K. Defence Science and Technology Laboratory, and in part by the Engineering and Physical Research Council (collaboration between U.S. DOD, U.K. MOD, and U.K. EPSRC through the Multidisciplinary University Research Initiative) under Grant EP/R018456/1.

\ifCLASSOPTIONcaptionsoff
  \newpage
\fi

\bibliographystyle{IEEEtran}
\bibliography{IEEEabrv,Semantic}

\end{document}